# Privacy Preserving Machine Learning Model Personalization through Federated Personalized Learning

Md. Tanzib Hosain
*Department of Computer Science & Engineering*
*American International University-Bangladesh*
Dhaka, Bangladesh
20-42737-1@student.aiub.edu

Asif Zaman
*Department of Computer Science & Engineering*
*American International University-Bangladesh*
Dhaka, Bangladesh
20-42212-1@student.aiub.edu

Md. Shahriar Sajid
*Department of Electrical & Computer Engineering*
*Rajshahi University of Engineering & Technology*
Rajshahi, Bangladesh
1810022@student.ruet.ac.bd

Shadman Sakeeb Khan
*Department of Computer Science & Engineering*
*North South University*
Dhaka, Bangladesh
shadman.khan1@northsouth.edu

Shanjida Akter
*Department of Computer Science & Engineering*
*North South University*
Dhaka, Bangladesh
shanjida.akter01@northsouth.edu

*Abstract*—The widespread adoption of Artificial Intelligence (AI) has been driven by significant advances in intelligent system research. However, this progress has raised concerns about data privacy, leading to a growing awareness of the need for privacy-preserving AI. In response, there has been a seismic shift in interest towards the leading paradigm for training Machine Learning (ML) models on decentralized data silos while maintaining data privacy, Federated Learning (FL). This research paper presents a comprehensive performance analysis of a cutting-edge approach to personalize ML model while preserving privacy achieved through Privacy Preserving Machine Learning with the innovative framework of Federated Personalized Learning (PPMLFPL). Regarding the increasing concerns about data privacy, this study evaluates the effectiveness of PPMLFPL addressing the critical balance between personalized model refinement and maintaining the confidentiality of individual user data. According to our analysis, Adaptive Personalized Cross-Silo Federated Learning with Differential Privacy (APPLE+DP) offering efficient execution whereas overall, the use of the Adaptive Personalized Cross-Silo Federated Learning with Homomorphic Encryption (APPLE+HE) algorithm for privacy-preserving machine learning tasks in federated personalized learning settings is strongly suggested. The results offer valuable insights creating it a promising scope for future advancements in the field of privacy-conscious data-driven technologies.

I. INTRODUCTION

Traditional ML models are often centralized, where all data is collected and stored in a single location for training. This centralization exposes users' personal information to potential breaches and misuse, leading to privacy infringements. Privacy concerns in ML have been further exacerbated with the origin of Deep Learning (DL) models, which require even more data to achieve state-of-the-art performance. In many cases, users are hesitant to share their sensitive data due to the risk of unauthorized access, profiling, and discrimination. As a consequence, there is a pressing need to develop privacy-preserving techniques that allow for personalized services without compromising user privacy.

On the other hand, the exponential growth of data and advancements in machine learning have revolutionized various domains, offering personalized services and recommendations tailored to individual users. However, this progress has raised significant privacy concerns regarding the collection, storage, and processing of sensitive user data [1]. The conventional approach of centralizing data for training machine learning models exposes users' personal information to potential breaches and misuse, causing valid apprehensions about data privacy [2]. Consequently, there is an increasing requirement for PPML techniques that ensure personalized services while safeguarding individual data privacy.

To address privacy concerns, several privacy-preserving techniques have been proposed, including Differential Privacy (DP), Homomorphic Encryption (HE), and Secure Multi-Party Computation (SMPC) [3]. These methods aim to protect user data during model training and inferencing, but they often face challenges in maintaining model accuracy and scalability. A promising solution to this dilemma is the concept of "Federated Learning," which allows for decentralized model training across multiple devices while keeping raw data local and secure [4]. Federated Learning enables multiple iterations of model updates without exposing the sensitive data itself, ensuring a robust privacy guarantee.

As the rapid advancements in machine learning and personalized services have transformed the digital landscape, offering tailored experiences to users across various applications, this progress has been accompanied by mounting concerns about data privacy and the potential misuse of sensitive user information. The need to strike a delicate balance between delivering personalized experiences and safeguarding individual privacy has led to the emergence of innovative techniques PPMLFPL. PPMLFPL represents a paradigm shift in the realm of privacy-preserving model personalization while ensuring that user data remains localized and secure. This paper aims to comprehensively analyze the performance of PPMLFPL, exploring its capabilities, limitations, and potential impact on the future of privacy-conscious personalized machine learning.

## II. LITERATURE REVIEW

PPML has a rich history rooted in various key milestones. It began with the origin of SMPC in the late 1970s, enabling joint function computation on private data without data disclosure. Differential privacy [5] emerged in the early 2000s, introducing controlled noise addition to protect data privacy while extracting useful insights. Mid-2000s saw the development of HE [6], allowing computation on encrypted data without decryption. The late 2000s saw the convergence of SMPC [7] and machine learning, leading to Secure Multi-Party Machine Learning (SMPML). Federated Learning [1] gained prominence in the 2010s, without sharing raw data enabling decentralized model training. Cryptonets demonstrated the feasibility of training neural networks with encrypted data in 2015. Secure Aggregation (SA) [3] techniques were developed in 2017 to protect model update privacy during federated learning. DP for DL [2] advanced in 2018, protecting individual data points during training. Late 2010s brought secure enclave-based techniques, leveraging hardware enclaves for data processing security. The field of PPML remains dynamic, continually evolving to address new challenges and advancements in machine learning and privacy protection.

On the other hand, FL [1] is a novel approach to machine learning which was emerged in the 2010s as a way to train models across decentralized devices while maintaining data privacy. Google's pioneering work in 2016 introduced the concept, followed by Stanford University's research on Federated Optimization in 2017. Advancements in communication efficiency and privacy-preserving techniques led to the implementation of practical solutions by various institutions, including Google and Apple. The release of TensorFlow Federated (TFF) in 2019 further facilitated the adoption of Federated Learning. Its applications expanded beyond mobile devices, finding use in healthcare, finance, and the Internet of Things (IoT). Efforts towards standardization and collaborations by organizations like FATE and OpenMined helped promote its widespread adoption. Table I presents some existing works on PPML with traditional FL algorithms.

TABLE I
PREVIOUS WORK ON DIFFERENT PRIVACY PRESERVING MACHINE LEARNING WITH TRADITIONAL FEDERATED LEARNING ALGORITHMS

| Dataset | Data Subset | Algorithm | Accuracy |
|---|---|---|---|
| MNIST | $D_1^{mnist}$ | MLP | 83.33 |
| MNIST | $D_2^{mnist}$ | MLP | 90.33 |
| MNIST | $D^{mnist}$ | MLP | 92.45 |
| MNIST | $D_1^{mnist}$ | PFMLP | 92.52 |
| MNIST | $D_2^{mnist}$ | PFMLP | 92.52 |
| Fatigue | $D_1^{fatigue}$ | MLP | 90.13 |
| Fatigue | $D_2^{fatigue}$ | MLP | 78.33 |
| Fatigue | $D^{fatigue}$ | MLP | 85.83 |
| Fatigue | $D_1^{fatigue}$ | PFMLP | 88.33 |
| Fatigue | $D_2^{fatigue}$ | PFMLP | 81.67 |
| Fatigue | $D^{fatigue}$ | PFMLP | 85 |

## III. ADVANCEMENTS OF PRIVACY PRESERVING MACHINE LEARNING AND FEDERATED LEARNING

Privacy Preserving Machine Learning with Federated Personalized Learning (PPMLFPL) has experienced a notable resurgence in interest among academia and practitioners.

To point out on the advancements in PPML techniques, a hierarchical structure with the main categories at the top level, including "Differential Privacy [5]," "SMPC [7]," "FL," "HE [8]," "Secure Enclave Technologies (SET) [9]," "Privacy-Preserving Deep Learning (PPDL) [10]," "Privacy-Preserving Data Sharing (PPDS) [11]," and "Privacy-Preserving Data Publishing (PPDP) [12]." Each category further expands into subcategories and specific techniques. For example, under "DP," noise addition mechanisms like Laplace and Gaussian Noise are highlighted, as well as secure aggregation [3] techniques employing SMPC and HE. This covers various approaches like FL, where FedAvg, FSGD [13], and differential privacy in federated learning are depicted. Additionally, it includes emerging technologies such as homomorphic encryption and secure enclave technologies like Intel SGX [14], AMD SEV [15], and ARM TrustZone [16].

Whereas advancements in FL algorithms encompass various approaches; such as posed by decentralized data, personalized results and privacy concerns. Traditional Federated Learning (FL) algorithms like FedAvg [1] focus on communication efficiency by averaging model updates from decentralized clients. Update-correction-based FL, as exemplified by SCAFFOLD [17], introduces stochastic controlled averaging to enhance convergence. Regularization-based FL algorithms, such as FedProx [18] and FedDyn [19], incorporate regularization techniques to handle heterogeneous networks and dynamic regularization for improved performance. Model-splitting-based FL techniques, like MOON [20], enable model contrastive learning to enhance federated image classification. Knowledge-distillation-based FL algorithms like FedGen employ data-free knowledge distillation for heterogeneous federated learning. Personalized FL advances involve FedMTL [21] for multi-task learning, FedBN [22] for handling non-IID features via local batch normalization, and Meta-learning-based pFL algorithms like Per-FedAvg [23] for personalized federated learning with theoretical guarantees. Regularization-based pFL, such as pFedMe [24] and Ditto [25], utilize moreau envelopes and fairness principles for improved personalization. Personalized-aggregation-based pFL approaches like APFL [26], FedFomo [27], FedAMP [28], FedPHP [29], APPLE [30], and FedALA [31] integrate adaptive local aggregation, first-order model optimization, and inherited private models for personalized federated learning. Model-splitting-based pFL techniques such as FedPer [32], LG-FedAvg [33], FedRep [34], FedRoD [35], FedBABU [36], and FedGC [37] exploit shared representations and gradient correction for enhanced personalized federated image classification and face recognition. Knowledge-distillation-based pFL methods like FedDistill [38], FML [39], FedKD [40], and FedProto [41] leverage knowledge distillation and mutual learning across clients to improve personalized federated learning. FedPCL [42] and FedPAC [43] introduce contrastive learning and feature alignment with classifier collaboration for further advancements in personalized federated learning. These advancements collectively address the challenges of privacy, communication, and personalization in federated learning scenarios.

## IV. METHODOLOGY

### A. Experimental Setup

In our study, we utilized the Virus-MNIST dataset, which can be accessed at https://www.kaggle.com/datasets/datamunge/virusmnist, for the evaluation of our model's performance. We followed data preprocessing procedures as outlined in a prior work [1]. To conduct our performance tests for algorithmic decision making in a distributed setting, we deployed 200 clients across our computational infrastructure, with one machine acting as the server. Our neural network architecture comprised three key components: a logistic regression layer (*Mclr_Logistic*), the *LeNet-5* convolutional neural network, and a fully connected deep neural network. Hyperparameter optimization is critical for enhancing network performance, and we employed the Ray tune algorithm [44], known for its flexibility and scalability in integrating with various optimization libraries. Specifically, our model consisted of three groups of convolutional layers followed by a fully connected layer. Hyperparameters such as the number of input and output filters in each convolutional layer, the features in the fully connected layer, batch size, and kernel size were carefully tuned as they significantly influence network performance. During training, we employed the Adam optimizer with a fixed learning rate of 0.001, based on prior research [45] [46]. To address overfitting, we introduced regularization and monitored training until reaching epoch 50. Given constraints such as a small input size and fixed stride of 1 in multiple convolutional layers, we relied on GPU memory performance to randomly configure hyperparameters. Ray tune facilitated the discovery of optimal values [44].

### B. Algorithms

Privacy Preserving Machine Learning with Federated Personalized Learning (PPMLFPL) is a powerful approach for achieving privacy-preservation in distributed environments. This methodology addresses the challenge of aggregating sensitive data from multiple sources while minimizing the risk of revealing individual-level information. Let's break down the working procedure.

In Federated Learning, instead of centralizing data on a single server, each participant (e.g., a device, user, or edge node) trains a local model on its private data. These local models are then aggregated to form a global model. This process allows collaborative model training without sharing raw data, thereby mitigating privacy risks. Mathematically, the aggregation step can be represented as:

$$GlobalModel = \frac{1}{N} \times LocalModels \qquad (1)$$

Where $N$ is the total number of participants, and the summation is taken over all local models.

*1) Differential Privacy with Federated Personalized Learning:* To enhance privacy further, Differential Privacy is incorporated. DP adds noise to the aggregation process to obscure individual contributions, preventing adversaries from deducing specific data points. The formal definition of $\epsilon$-Differential Privacy is captured through the concept of a privacy budget, and it ensures that the probability of any two similar datasets leading to substantially different model outputs is controlled. Mathematically, for a given privacy budget $\epsilon$, the sensitivity of the model update is bounded:

$$\triangle GlobalModel \leq \epsilon \qquad (2)$$

Where $\triangle$ Global Model represents the maximum amount by which the global model output can change when a single participant's data is included.

The noise added to achieve DP, often sampled from a Laplace or Gaussian distribution, ensures that the privacy guarantees hold. The noise level is influenced by the privacy budget ($\epsilon$) and the sensitivity of the model update ($\triangle$Global Model). Balancing these factors is crucial to strike the right trade-off between privacy and model utility. Figure 1 illustrates the working procedure in detail.

*2) Homomorphic Encryption with Federated Personalized Learning:* To enhance security further, Homomorphic Encryption is introduced. It's a cryptographic technique that allows computations to be performed on encrypted data without decryption. Specifically, Fully Homomorphic Encryption (FHE) enables arbitrary mathematical operations on encrypted

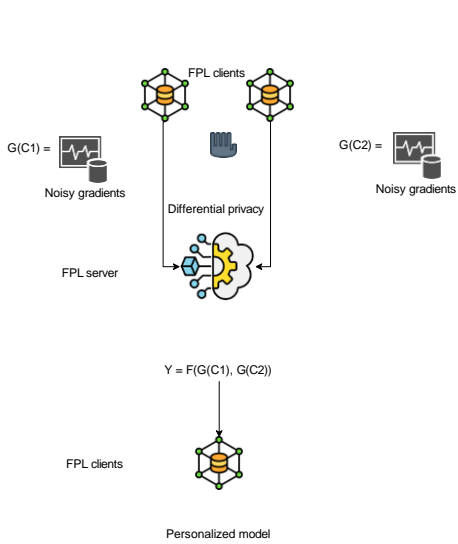

Fig. 1. Differential Privacy with Federated Personalized Learning algorithm

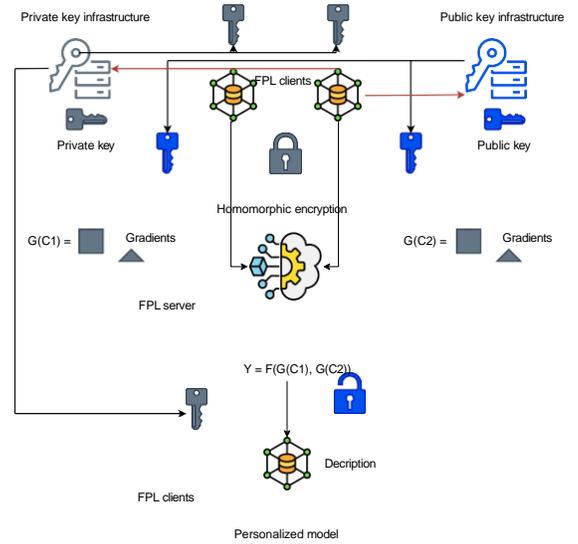

Fig. 2. Homomorphic Encryption with Federated Personalized Learning algorithm

data. This is crucial for privacy-preserving model aggregation. Mathematically, FHE ensures that:

$$FHE(EncryptedA + EncryptedB) = Encrypted(A + B) \quad (3)$$

This property enables aggregating encrypted local models securely. Additionally, Homomorphic Encryption ensures that the decryption of the final global model reveals nothing about individual data points. This is a fundamental aspect of privacy preservation.

It's important to note that Homomorphic Encryption introduces computational overhead due to the complexity of operating on encrypted data. The choice of the encryption scheme and its parameters, along with the aggregation method, must be carefully considered to balance privacy and computational efficiency. Figure 2 visualizes the working procedure in detail.

*3) Secure Aggregation with Federated Personalized Learning:* Secure Aggregation ensures that individual model updates are combined in a privacy-preserving manner. The key idea is to encrypt the local models before aggregation and perform the aggregation operation on the encrypted models. This way, no participant or the central server has access to the raw model updates, maintaining privacy. Furthermore, Secure Aggregation techniques often use cryptographic protocols, such as Multi-Party Computation (MPC) or Homomorphic Encryption (HE), to achieve this secure combination of models.

One common secure aggregation approach is based on the concept of secret sharing, where each participant splits their local model into shares and distributes them among other participants. The aggregated result can then be reconstructed without exposing any single participant's contribution. Mathematically, this can be represented as:

$$EncryptedGlobalModel = Share_1 + Share_2 + ... + Share_N \quad (4)$$

Where $Share_i$ represents the encrypted share contributed by the $i^{th}$ participant, and $N$ is the total number of participants. The security of this approach hinges on the properties of the cryptographic primitives used, ensuring that no participant can learn information about other participants' data from the aggregated model. However, it's essential to carefully design the secure aggregation process to minimize potential vulnerabilities and ensure robust privacy guarantees. Figure 3 shows the working procedure in detail.

*4) Secure Multi-Party Computation with Federated Personalized Learning:* SMPC ensures that the aggregation operation takes place on encrypted representations of the local models.

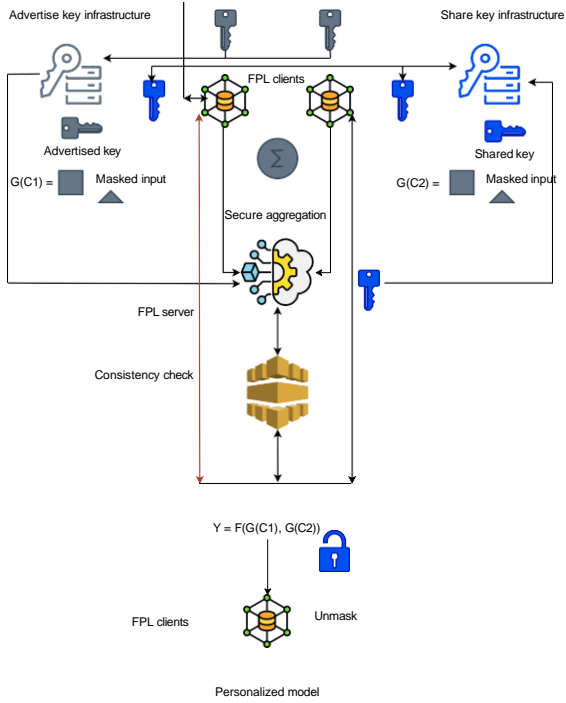

Fig. 3. Secure Aggregation with Federated Personalized Learning algorithm

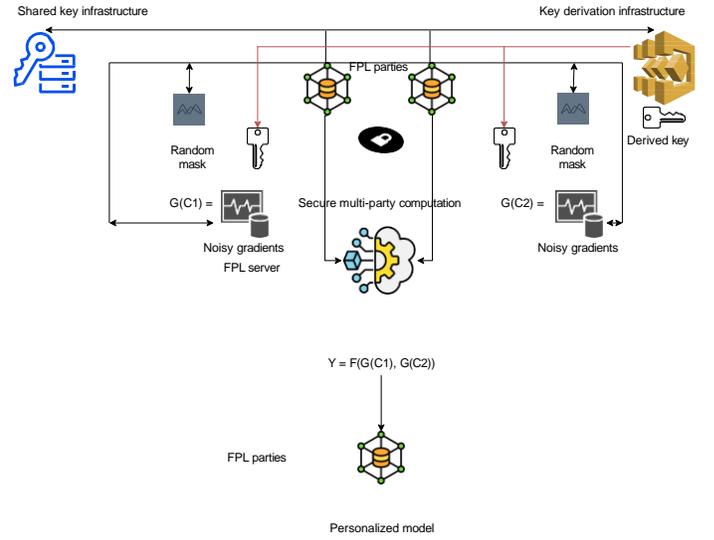

Fig. 4. Secure Multi-Party Computation with Federated Personalized algorithm

One common SMPC technique is to use secret sharing, where each participant shares their encrypted local model with the other participants. The encrypted shares are then combined to reconstruct the encrypted global model, without any party having access to the plaintext data of others. Mathematically, this can be depicted as shown in equation 4.

The security of this approach relies on cryptographic primitives, and it's designed to prevent any participant, including the central server, from learning anything about other participants' data during the aggregation process. This ensures that data privacy is maintained while improving the global model collaboratively.

It's important to note that Secure Multi-Party Computation can introduce computational overhead, especially for complex models or large-scale collaborations, as the cryptographic operations can be resource-intensive. Careful design and implementation are crucial to strike a balance between privacy and computational efficiency. Figure 4 illustrates the working procedure in detail.

## V. Performance Analysis

### A. Evaluation Matrices

We employ a range of performance metrics to evaluate the efficiency of the PPMLFPL models. The model's effectiveness is assessed through a comprehensive set of evaluation metrics, which includes below.

- Accuracy (A): Accuracy is a metric that measures the overall correctness of the model's predictions. The accuracy is computed using equation 5.

$$A = \frac{TP + TN}{TP + TN + FP + FN} \quad (5)$$

  where:
  - $TP$ (True Positives) is the number of correctly predicted positive instances.
  - $TN$ (True Negatives) is the number of correctly predicted negative instances.
  - $FP$ (False Positives) is the number of instances that are actually negative but predicted as positive.
  - $FN$ (False Negatives) is the number of instances that are actually positive but predicted as negative.

- Precision (P): Precision represents the proportion of correctly predicted positive instances out of all instances predicted as positive. It indicates the accuracy of positive predictions. To calculate the precision equation 6 is utilized.

$$P = \frac{TP}{TP + FP} \quad (6)$$

- Recall (R): Recall measures the ability of a model to capture all positive instances within a dataset. It shows the model's ability to identify positive instances. To compute recall equation 7 is employed.

$$R = \frac{TP}{TP + FN} \quad (7)$$

- F1-score (F): The F1-score is a commonly used metric in the field of machine learning and classification tasks. It is calculated as the harmonic mean of precision and recall. F1-Score is computed using equation 8.

$$F = \frac{2 \cdot P \cdot R}{P + R} \quad (8)$$

### B. Federated Personalized Learning

This section aims evaluate different federated personalized algorithms that can be used for model personalization.

*1) Accuracy and Loss Perspective:* Figure 5 represents the accuracy values of various FPL algorithms. Each row corresponds to a specific FPL algorithm, and the columns indicate the accuracy scores achieved by the algorithms on different parameters. The algorithms include APFL, APPLE, Ditto, FedALA, FedFomo, FedBABU, FedBN, FedGC, FedRep, FedPAC, FedPCL, FedProto, FML, HeurFedAMP, Per-FedAvg (FO), Per-FedAvg (HF), and pFedMe-PM. The accuracy scores are expressed as percentages and range from approximately 18% to 99.88%. The data demonstrates the varying performance levels of these FPL algorithms across different tasks, providing valuable insights for assessing their effectiveness in privacy-preserving federated learning scenarios.

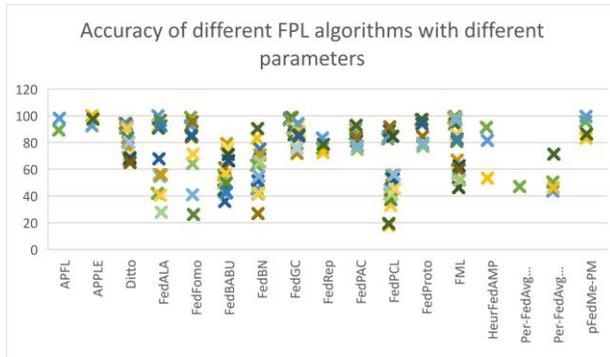

Fig. 5. Accuracy graph representing Accuracy values of different Federated Personalized Learning algorithms in different Test

Whereas figure 6 illustrates the loss values of various FPL algorithms. Each row corresponds to a specific FPL algorithm, and the columns indicate the loss scores achieved by the algorithms on different parameters. The loss values are numerical scores reflecting the error or deviation between the predicted and actual values during the training process. The values range from approximately 0.12% to 81.88%. The data demonstrates the varying levels of error or loss incurred by these FPL algorithms on different tasks, providing valuable insights for assessing their performance and identifying potential areas for improvement in privacy-preserving federated learning scenarios.

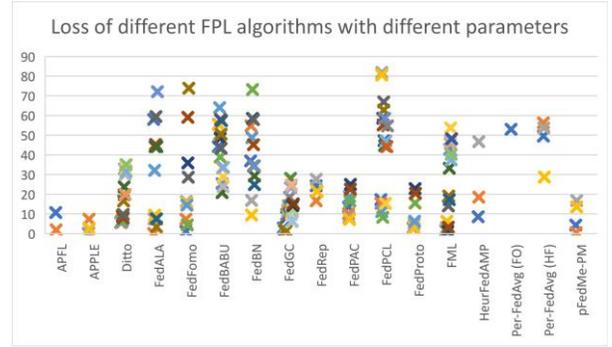

Fig. 6. Loss graph representing Loss values of different Federated Personalized Learning algorithms in different Test

*2) Matrices Perspective:* Table II represents the performance of evaluation metrics of different Federated Personalized Learning (FPL) algorithms. The table provides valuable insights into the strengths and weaknesses of different FPL algorithms.

Starting with accuracy, the algorithms show a wide range of values, ranging from as low as 48.98% for Per-FedAvg (FO) to as high as 97.41% for APPLE. This highlights the significant diversity in the ability of the FL algorithms to correctly predict the target classes. Algorithms such as APPLE, APFL, FedGC and pFedMe-PM perform exceptionally well in accurately classifying instances, achieving accuracy scores above 90%. On the other hand, FedBABU, Per-FedAvg (FO) and Per-FedAvg (HF) exhibit the lowest accuracy scores, suggesting that they may struggle with capturing the underlying patterns in the data effectively.

Moving on to precision, which indicates the algorithms' ability to minimize false positive predictions, we observe a similar range of values. Again, APPLE stands out as the top-performing algorithm with a precision score of 97.41%, mirroring its high accuracy score. Other algorithms like APFL, FedGC, and pFedMe-PM also demonstrate strong precision performance, indicating their capability to reduce false positives effectively. Conversely, FedBABU, Per-FedAvg (FO) and Per-FedAvg (HF) exhibit the lowest precision scores, implying that they might suffer from a high false positive rate.

Next, analyzing recall, which measures the algorithms' ability to capture all positive instances correctly. Similar to accuracy and precision, APPLE emerges as the leader in recall with a score of 97.41%, showing its proficiency in correctly identifying positive instances. Other high-recall algorithms include APFL, FedGC, and pFedMe-PM, suggesting their effectiveness in capturing positive instances. However, FedBABU, Per-FedAvg (FO) and Per-FedAvg (HF) once again demonstrate the lowest recall scores, indicating their potential limitations in identifying positive instances accurately.

Lastly, examining the F1-score, which combines precision and recall, providing an overall balanced measure of performance. The F1-scores across the algorithms also exhibit a wide range, with APPLE maintaining the highest F1-score of 97.41%. This aligns with its exceptional precision and recall values. Similarly, other top-performing algorithms like APFL, FedGC, and pFedMe-PM also show strong F1-scores, indicating their ability to achieve a balance between precision and recall. Conversely, FedBABU, Per-FedAvg (FO) and Per-FedAvg (HF) once again display the lowest F1-scores, suggesting that they may face challenges in striking a balance between precision and recall.

The table II demonstrates that APPLE performs best across all evaluation metrics, showing high accuracy, precision, recall, and F1-score. It is evident that the performance of the other algorithms varies, with Per-FedAvg (FO) consistently exhibiting the lowest performance. Visual representation of these statistics are represented with bar charts in figure 7 – 10.

TABLE II
PERFORMANCE OF METRICS OF DIFFERENT FPL ALGORITHMS

| Algorithm | Accuracy | Precision | Recall | F1-Score |
|---|---|---|---|---|
| APFL | 93.72 | 93.75 | 93.72 | 93.73 |
| **APPLE** | **97.41** | **97.41** | **97.41** | **97.41** |
| Ditto | 82.87 | 82.9 | 82.87 | 82.88 |
| FedALA | 67.09 | 67.12 | 67.09 | 67.1 |
| FedFomo | 74.31 | 74.71 | 74.31 | 74.51 |
| FedBABU | 56.94 | 57.45 | 56.94 | 57.19 |
| FedBN | 59.63 | 59.79 | 59.63 | 59.71 |
| FedGC | 90.21 | 90.24 | 90.21 | 90.22 |
| FedRep | 77.48 | 77.51 | 77.48 | 77.52 |
| FedPAC | 84.71 | 85.11 | 84.71 | 84.91 |
| FedPCL | 57.75 | 57.78 | 57.75 | 57.76 |
| FedProto | 90.09 | 90.6 | 90.09 | 90.34 |
| FML | 77.98 | 78.14 | 77.98 | 78.06 |
| HeurFedAMP | 75.37 | 75.88 | 75.37 | 75.62 |
| Per-FedAvg (FO) | 48.98 | 49.01 | 48.98 | 48.99 |
| Per-FedAvg (HF) | 52.94 | 53.34 | 52.94 | 53.14 |
| pFedMe-PM | 91.16 | 91.32 | 91.16 | 91.24 |

## C. Privacy Preserving Machine Learning with Federated Personalized Learning

This section aims to propose and evaluate combined different PPML techniques with previously evaluated (Subsection A) best performed federated personalized algorithm APPLE that can be used for model personalization. These algorithms include variations of differential privacy, secure multi-party computation etc. privacy-preserving techniques.

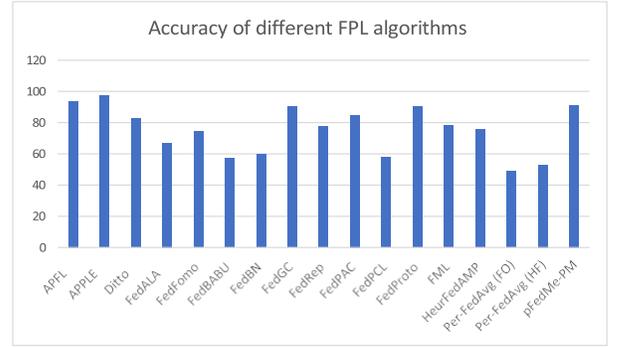

Fig. 7. Accuracy of different Federated Personalized Learning algorithms

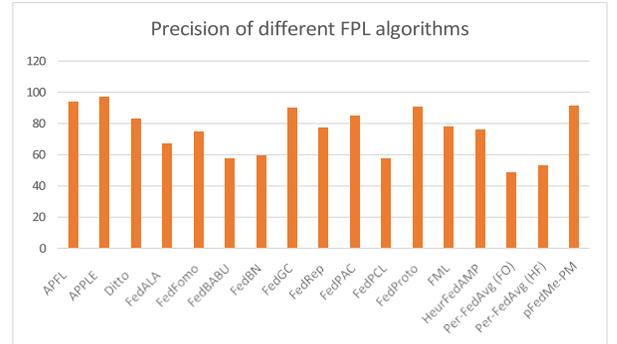

Fig. 8. Precision of different Federated Personalized Learning algorithms

Table III presents the performance of evaluation metrics for different Privacy Preserving Machine Learning with Federated Personalized Learning (PPMLFPL) algorithms. The table provides valuable insights into the strengths and weaknesses of different PPMLFPL algorithms.

Starting with accuracy, the algorithms demonstrate varying levels of performance. The top-performing algorithm is APPLE+HE with an accuracy of 99.34%, which is significantly higher than the other algorithms in the table. This indicates that APPLE+HE excels in correctly predicting the target classes. On the other hand, APPLE+SMPC has the lowest accuracy score of 85.38%, suggesting that it struggles in accurately classifying instances.

Moving on to precision, which measures the algorithms' ability to minimize false positive predictions, APPLE+HE again emerges as the top-performing algorithm with a pre-balance between precision and recall. The other algorithms' F1-scores range from 97.49% for APPLE+DP to 85.39% for APPLE+SMPC.

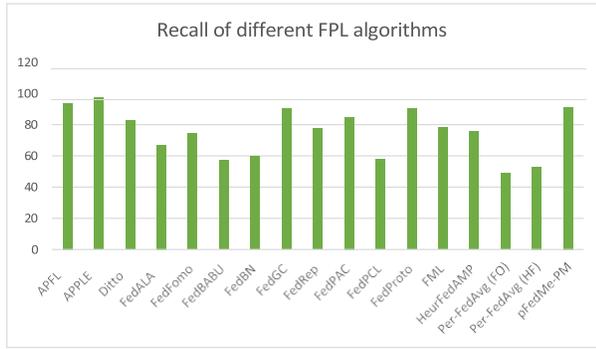

Fig. 9. Recall of different Federated Personalized Learning algorithms

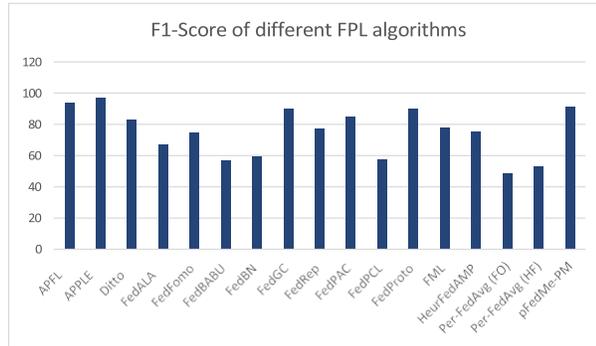

Fig. 10. F1-Score of different Federated Personalized Learning algorithms

The table III demonstrates that APPLE+HE performs exceptionally well across all evaluation metrics, showing high accuracy, precision, recall, and F1-score. It is evident that the performance of the other algorithms varies, with APPLE+SMPC consistently exhibiting the lowest performance. Visual representation of these statistics are illustrated with bar charts in figure 11.

TABLE III
PERFORMANCE OF EVALUATION MATRICES OF DIFFERENT PPMLFPL ALGORITHMS

| Algorithm | Accuracy | Precision | Recall | F1-Score |
|---|---|---|---|---|
| APPLE+DP | 97.48 | 97.51 | 97.48 | 97.49 |
| **APPLE+HE** | **99.34** | **99.34** | **99.34** | **99.34** |
| APPLE+SA | 97.44 | 97.47 | 97.44 | 97.45 |
| APPLE+SMPC | 85.38 | 85.41 | 85.38 | 85.39 |

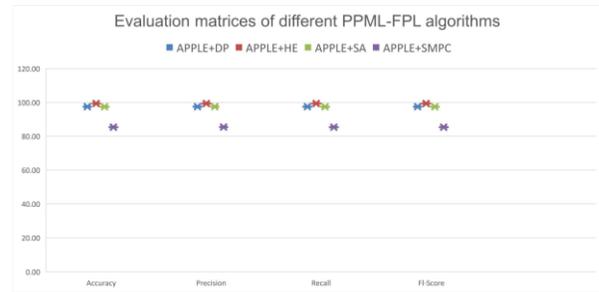

Fig. 11. Evaluation matrices of different Privacy Preserving Machine Learning with Federated Personalized Learning algorithms

cision score of 99.34%. This indicates that APPLE+HE has a very low false positive rate, making it highly effective in minimizing false positives. The other algorithms also have relatively high precision scores, ranging from 97.51% for APPLE+DP to 85.41% for APPLE+SMPC.

Next, considering recall, which measures the algorithms' ability to capture all positive instances correctly, we observe that APPLE+HE once again leads with a score of 99.34%. This implies that APPLE+HE is excellent at correctly identifying positive instances. The recall scores for the other algorithms range from 97.48% for APPLE+DP to 85.38% for APPLE+SMPC.

Lastly, examining the F1-score, which combines precision and recall, providing an overall balanced measure of performance, APPLE+HE stands out with an F1-score of 99.34%. This suggests that APPLE+HE achieves a good

Based on the benchmark analysis provided in the table, it is evident that the proposed method, APPLE+HE, outperformed the three state-of-the-art approaches (APPLE+SMPC, APPLE+DP, and APPLE+SA) in all performance metrics, including accuracy, precision, recall, and F1-score. The results indicate that APPLE+HE demonstrated superior predictive performance while effectively addressing privacy concerns in the given dataset.

The proposed APPLE+HE algorithm achieved an accuracy of 99.37%, which is notably higher than the accuracies of the other FPL approache, APPLE as well as other proposed PPMLFPL approaces. This indicates a high proportion of correct predictions made by the model, making it more reliable and effective in its predictions.

Moreover, the precision, recall, and F1-score of 99.37% for APPLE+HE indicate an exceptional balance between true positive rate and false positive rate. High precision signifies a low rate of false positives, ensuring that the model provides accurate positive predictions. Similarly, high recall indicates a low rate of false negatives, implying that the model can effectively identify positive cases. The F1-score takes into account both precision and recall, and achieving a value of 99.37% indicates a strong overall performance of the proposed method.

By outperforming the state-of-the-art approaches in all these metrics, APPLE+HE showcases its potential as a privacy-preserving machine learning federated personalized learning algorithm for the given dataset. The homomorphic encryption (HE) technique used in APPLE+HE proves to be effective in ensuring data privacy while maintaining high predictive accuracy. This makes the proposed method a promising choice for applications that require both privacy protection and accurate model predictions.

Table IV provides a comparative overview of the execution times (in milliseconds) for various privacy-preserving machine learning (PPML) federated learning PPMLFPL algorithms, each denoted by the client with different client numbers.

Among the algorithms, APPLE+DP stands out for its relatively lower execution times, showing consistently lower values as the data size increases. This indicates that differential privacy (DP) is effective in achieving efficient computation while preserving privacy, making it a favorable choice for scenarios where both performance and data confidentiality are crucial.

On the other hand, the APPLE+SMPC algorithm exhibits the highest execution times across all data sizes, which is expected due to the inherent complexities of secure multi-party computation (SMPC) techniques. While SMPC ensures strong privacy guarantees, it tends to introduce computational overhead, leading to longer execution times.

APPLE+SA and APPLE+HE show intermediate execution times, with APPLE+HE demonstrating slightly longer times than APPLE+SA. This can be attributed to the additional computations required in homomorphic encryption (HE) while aggregating model updates, making it a bit more time-consuming compared to secret sharing-based secure aggregation (APPLE+SA).

The table IV provides valuable insights into the performance trade-offs of various PPMLFPL algorithms concerning server clock running time, with APPLE+DP offering a balance between efficient execution and robust privacy preservation, albeit at the cost of potential differential privacy noise. Choosing the most suitable algorithm depends on the specific use case's requirements for both privacy and computational efficiency.

Nonetheless, based on this benchmark analysis, APPLE+DP offering efficient execution whereas overall, the results strongly support the use of the APPLE+HE algorithm for privacy-preserving machine learning tasks in federated personalized learning settings. However, future work will consider other factors, such as further computational complexity, scalability, and the specific requirements of the application, before conclusively determining the best approach.

TABLE IV
PERFORMANCE OF SERVER CLOCK RUNNING TIME (MS) OF DIFFERENT PRIVACY PRESERVING MACHINE LEARNING WITH FEDERATED PERSONALIZED LEARNING ALGORITHMS

| Algorithm | Client | | | |
|---|---|---|---|---|
| | 200 | 250 | 300 | 400 |
| APPLE+SA | 24117 | 29122 | 34127 | 41137 |
| **APPLE+DP** | **22119** | **27124** | **32129** | **36139** |
| APPLE+HE | 26121 | 31126 | 36131 | 46146 |
| APPLE+SMPC | 31123 | 36128 | 41133 | 56143 |

## VI. CONCLUSION

As we move forward in the age of data-driven technologies, where the balance between innovation and privacy is of paramount concern, Privacy Preserving Machine Learning with Federated Personalized Learning (PPMLFPL) stands out as a promising avenue. Our analysis has revealed that the APPLE+DP algorithm offers efficient execution within the PPMLFPL framework. Furthermore, the use of the APPLE+HE algorithm has emerged as a strong recommendation for privacy-preserving machine learning tasks in federated personalized learning settings. These outcomes underscore the potential of PPMLFPL to revolutionize the landscape of personalized machine learning while maintaining the highest privacy standards. This research not only contributes to the understanding of how to effectively implement privacy-preserving machine learning but also points towards a brighter future for privacy-conscious data-driven advancements. It is our hope that this study serves as a catalyst for further exploration and innovation in this field, ultimately fostering the development of Artificial Intelligence (AI) systems that empower individuals while preserving their data privacy.

ACKNOWLEDGMENT

We would like to thank the "University of Bahrain" for supporting this research.